\documentclass{icis}
\usepackage{authblk} 
\usepackage{graphicx}
\usepackage{xcolor}
\usepackage{amsmath}
\usepackage{tabularray} 
\usepackage{subfigure}
\title{Prediction of Bank Credit Ratings using \\ Heterogeneous Topological Graph Neural Networks}
\researchtype{Regular Paper}
\shorttitle{Bank Credit Rating Prediction with Graph Neural Networks}
\conference{Workshop on Information Technologies and Systems, Bangkok, Thailand 2024}

\usepackage[hidelinks]{hyperref}

\usepackage{tabularx}  
\usepackage{subfigure}
\usepackage{multirow}

\usepackage{dsfont}

\begin{document}

\maketitle

    

\bgroup
\begin{table}[!tbh]
\begin{tblr}{
colspec={Q[c,wd=0.45\textwidth] Q[c,wd=0.45\textwidth]},
width=\textwidth,
rowsep = 1pt,    
rows = {font=\fontsize{13pt}{15pt}\selectfont} 
}
\textbf{Junyi Liu}   &  \textbf{Stanley Kok}    \\
 Department of Information Systems and Analytics
 & Department of Information Systems and Analytics \\
National University of Singapore & National University of Singapore \\
 junyiliu@u.nus.edu & skok@comp.nus.edu.sg
    \end{tblr}
    \caption*{}
\end{table}
\vspace{-0.3in}
\egroup





  

\section{Introduction} 


Bank credit ratings, assigned by agencies like Standard \& Poor's, Moody's, and Fitch, evaluate a bank's financial health based on factors such as asset quality, profitability, and market position~\autocite{white2010markets}. These ratings are critical indicators of a bank's ability to repay debt and significantly influence economic players: for businesses, they affect borrowing costs and market trust; for economies, they impact financial system stability. Sudden rating changes can trigger volatile capital flows and market fluctuations, influencing economic growth and financial stability.

During financial market instability, predicting bank credit ratings, especially for the upcoming quarter, becomes crucial. These predictions provide the data needed for informed decision-making, prompt regulatory adjustments, and the protection of investors and the public. The 2023 bankruptcy of Silicon Valley Bank (SVB), which triggered collapses like those of Signature Bank and First Republic Bank, underscores the resulting financial turmoil~\autocite{aharon2023too}. 


Graph neural networks (GNNs) have become a pivotal technology in financial risk prediction, particularly excelling in node classification and link prediction tasks~\autocite{wu2022graph}. These models effectively leverage edge information to represent the propagation of financial risk within networks. However, the full graph of interbank connections is often unavailable due to the private nature of these relationships, posing a challenge for directly applying GNNs to predict bank credit ratings.

To address this, researchers infer networks from interbank lending records, where each bank is represented as a node, and an edge connects two banks if a lending relationship exists~\autocite{anand2015filling}. These relationships reflect a bank's liquidity and credit risk, making them useful for predicting credit ratings. Financially sound banks typically prefer to lend to each other, avoiding riskier institutions to enhance credit ratings and market stability by maintaining high capital ratios and choosing low-risk borrowers~\autocite{subramanyam2014financial}. While this homophilic tendency highlights similarities among banks with similar credit ratings, it only partially captures a bank's creditworthiness relative to its network neighbors~\autocite{ruddy2021analysis}.

We address this limitation by constructing an interbank lending network using a real-world dataset (detailed in our model description). Our analysis shows that only 25\% of connected banks share the same credit rating, indicating low homophily within the network. The remaining three-quarters of connections link banks with differing ratings, suggesting that this heterogeneity may hinder accurate credit rating predictions. To enhance node homogeneity and enrich our predictions, we generate topological features from bank statements. These statements are converted into multi-dimensional arrays containing data such as assets, liabilities, and deposits. We model banks as points in a high-dimensional space, where each point represents a bank's financial profile. Initially, these points lack topological structure, but using persistent homology~\autocite{wasserman2018topological}, we can observe how connections form and evolve into a complex network. This approach identifies enduring structural features that reveal stability and significance throughout the network's evolution~\autocite{otter2017roadmap}.



In persistent homology, we capture spatial relationships between points (banks) using the {\it Rips complex}. Unlike methods requiring geometric coordinates, the Rips complex focuses on distances between multi-dimensional points, making it ideal for high-dimensional data like bank statements~\autocite{attali2011vietoris}. The Rips complex connects points within a certain distance threshold, while persistent homology analyzes how these connections and resulting structures evolve as the threshold increases. By identifying structures that persist across thresholds, we uncover robust, meaningful relationships among banks, which provide valuable insights for graph neural networks, leading to more accurate bank credit rating predictions.


In this paper, we introduce the Heterogeneous Topological Graph Neural Network (HTGNN), a novel model for predicting bank credit ratings. HTGNN utilizes persistent homology to construct a network that captures relationships among banks and combines this with a traditional lending network to create a heterogeneous network that integrates information from both sources, leading to improved predictions. To our knowledge, this is the first application of persistent homology with GNNs for bank credit rating prediction. We validate the effectiveness of HTGNN through experiments on a real-world global-scale bank dataset.



\section{Related Work} 

 \subsection{Financial Risk Prediction} 

 Financial risk prediction is vital for maintaining a stable financial system and promoting long-term economic growth by forecasting events like bankruptcies and serving as an early warning system~\autocite{bonsall2017managerial}. This enables stakeholders to address vulnerabilities in financial entities before they escalate. A key task is predicting bank credit ratings, which assess creditworthiness and influence lending relationships and risk management strategies~\autocite{brockman2024credit}. 

Traditionally, financial risk prediction relied on accounting-based methods~\autocite{ohlson1980financial}. Recently, machine learning (ML) has offered more effective solutions. Network-based ML methods using graph theory have shown promise in analyzing default risk~\autocite{billio2012econometric, das2019matrix}, but often overlook the network's structure. Graph neural networks (GNNs) have emerged as powerful tools for financial risk prediction. They've been used effectively for predicting SME bankruptcies, building company-specific risk graphs, and enhancing credit rating predictions.  However, these studies primarily focus on enterprises and use graphs that are easily derived from available information. In contrast, our HTGNN model addresses the challenge of predicting bank credit ratings without disclosed bilateral relationships and with an incomplete network structure.

\subsection{Topology Data Analysis in the Financial Domain} 


Topological data analysis (TDA) is a branch of mathematics that reveals hidden structures in data, offering advantages over traditional network analysis techniques, such as robustness to noise, the ability to capture global and multi-scale features, and applicability to high-dimensional data. In this work, we focus on a specific TDA method called {\it persistent homology} (PH), which applies algebraic topology to analyze complex, noisy datasets, even those with many dimensions~\autocite{wasserman2018topological}.


The field of information systems has yet to fully embrace TDA techniques. Although some initial forays exist, TDA has not seen widespread adoption. For example, Ando et al. (2022) introduced a novel approach using quantile regression and vector autoregression to analyze how network topology changes under varying financial shocks. However, their work mainly focuses on understanding these topological shifts rather than solving specific problems.

Persistent homology has seen limited exploration in finance literature, primarily for quantifying transient cycles in multi-scale time-series data to estimate market performance~\autocite{ismail2022early}. These studies mostly use persistent homology to construct comparative indicators, rather than fully leveraging its network analysis capabilities. In this context, our HTGNN model pioneers the application of persistent homology to financial network analysis for bank credit rating prediction. 

\section{Background} 


\subsection{Persistent Homology and Associated Topological Concepts}

Persistent homology, a core tool in topological data analysis (TDA), is well-suited for analyzing complex, high-dimensional, and noisy datasets by identifying stable topological structures across scales. In finance, this enables the detection of enduring relationships between banks within intricate transaction networks, even in the presence of noise. 

Directly analyzing high-dimensional continuous spaces, such as bank statement, can be computationally expensive. To mitigate this, we approximate the space using a simplicial complex, specifically the Rips complex, which is a collection of $k$-simplices (e.g., points, edges, triangles) built from simpler shapes. A $k$-simplex is defined by $k$ points $\{x_1,\ldots,x_k\}$ within a distance threshold $r \in \mathds{R}$, i.e., $|| x_i-x_j || \leq r$. The dimension of the Rips complex is the largest $k$-simplex it contains.

As the scale $r$ increases, network components merge, new loops form, and voids disappear, making the network more connected. These changes are captured by the homology groups $H_m$, where $H_0$ includes connected components, $H_1$ captures loops, and $H_2$ identifies voids. For our HTGNN model, we focus on homology groups of dimensions $0 \leq m \leq 2$.

By varying the scale $r$ from $r_0$ to $r_{\max}$, we generate a filtration of homology groups, where each group at a smaller scale is contained within the group at a larger scale, i.e., $H_m(R(X, r_0)) \subseteq H_m(R(X, r_1)) \subseteq \ldots \subseteq H_m(R(X, r_{\text{max}}))$. As $r$ increases, the number of holes decreases or remains constant. Figure 1 illustrates the filtration of a Rips complex as the scale is increased. 
\vspace{-0.2cm} 
\begin{figure}[h] 
	\[ 
	\begin{array}{|c|} 
	\hline \\ [-11pt] 
	\includegraphics[scale = 0.2]{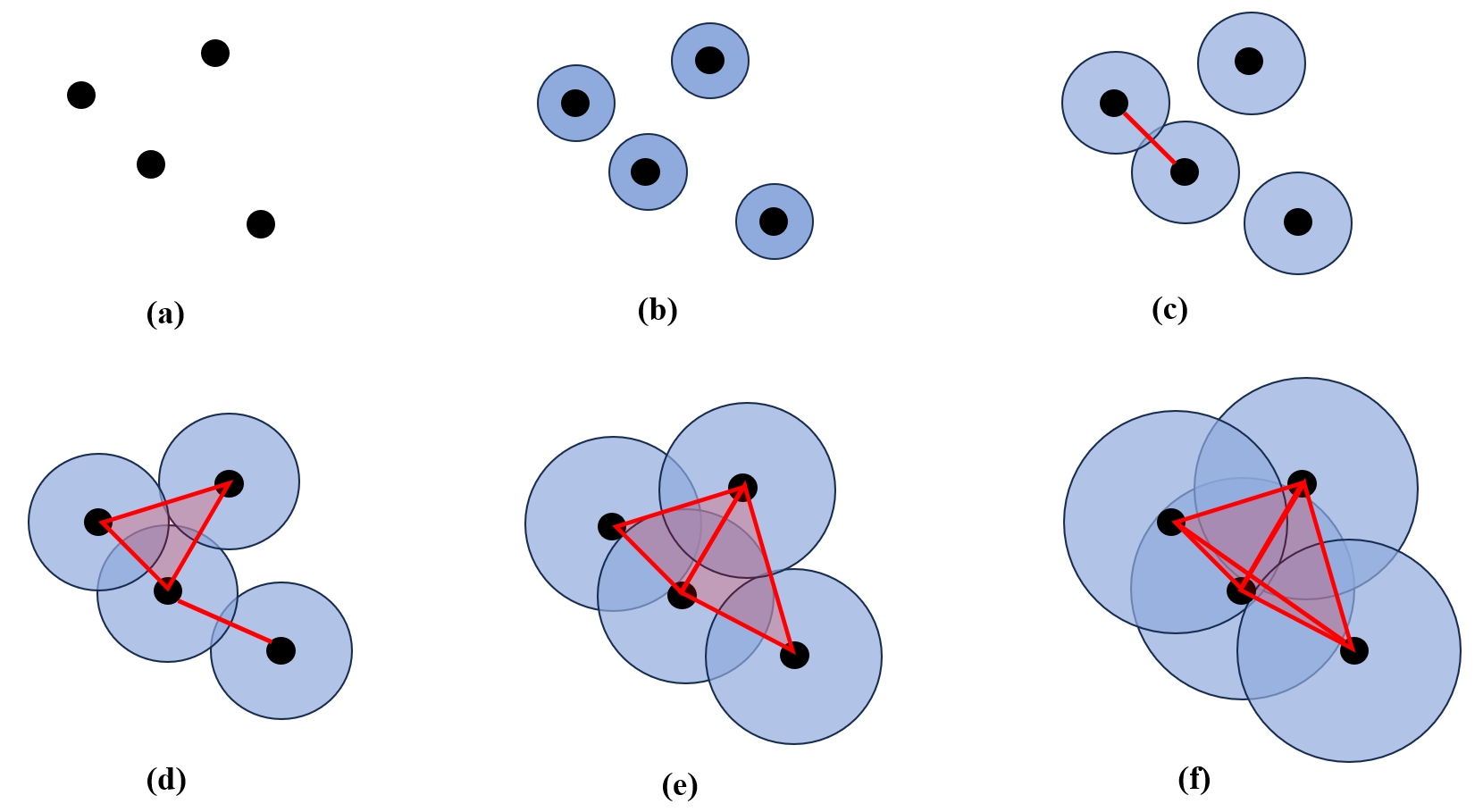} \\ [-4pt] 
	\hline 
	\addstackgap[2.5pt]{{\usefont{T1}{ptm}{b}{n}Figure 1.\hspace{0.09cm} A filtration with six Rips complexes built at different scales. Four connected }}\\ 
	\addstackgap[2.5pt]{{\usefont{T1}{ptm}{b}{n} components appear in (a) and (b), three appear in (c), one connected component and }}\\ 
   	\addstackgap[2.5pt]{{\usefont{T1}{ptm}{b}{n} one loop appear in (d), and higher-dimensional connected components appear in (e) and (f).}}\\ 
	\hline 
	\end{array} 
	\] 
\end{figure} 
\vspace{-0.9cm} 

Persistent homology tracks these topological features across scales, identifying the ranges $(r_b, r_d)$ where features appear (birth) and disappear (death). The difference $r_d - r_b$ indicates the feature's persistence, highlighting significant, long-lasting structures amidst the noise.

\subsection{Graph Neural Network}

Graphs are data structures used in machine learning to represent networks of interconnected objects (nodes) and the relationships between them (edges). Graph neural networks (GNNs) have become increasingly popular for graph-related tasks in recent literature~\autocite{wu2020comprehensive}. We next examine the information propagation in a GCN (Kipf and Welling 2016) model.



 A graph with $N$ nodes is represented by an $N \times N$ adjacency matrix $A$, where $A_{ij} = 1$ if an edge exists between nodes $i$ and $j$, and 0 otherwise. To allow nodes to propagate information to themselves, GCN adds self-loops by defining $\tilde{A} = A + I$, where $I$ is the identity matrix. Each node is associated with a $d$-dimensional vector, and all vectors are stacked in the $N \times d$ embedding matrix $H^{(l)}$, where $l$ denotes the propagation step. The transformation from $H^{(l)}$ to $H^{(l+1)}$ is governed by a $d \times d$ weight matrix $W^{(l)}$, which captures information propagation across edges. To prevent nodes with higher degrees from dominating, GCN normalizes the influence of each node using a diagonal matrix $\tilde{D}$, where $\tilde{D}_{ii}$ is the sum of the $i$-th row in $\tilde{A}$. The propagation from $H^{(l)}$ to $H^{(l+1)}$ is expressed as: $
H^{(l+1)} = \sigma(\tilde{D}^{-\frac{1}{2}} \tilde{A} \tilde{D}^{-\frac{1}{2}} H^{(l)} W^{(l)})
$, where $\sigma$ is a pointwise activation function like $ReLU$~\autocite{nair2010rectified}. The initial embedding $H^{(0)} = X$ represents the $d$-dimensional features of the $N$ nodes. The final embedding $H^{(L)}$ is used to predict task-specific values.

\section{Problem Formulation} 

In this paper, we use bank statement data to construct a network. We define $N$ nodes $V = \{v_1, \ldots, v_N\}$, where each node $v_i$ represents a bank and is associated with a feature vector $x_i$. These features form the node feature matrix $X = [x_1, \ldots, x_N]^T \in R^{N \times d}$, which serves as the initial layer $H^0$. We aim to learn the network structure of banks from their features (e.g., liquidity, income, and amount of deposits). Using $X$, we define a function $E: V \times V \to \{0, 1\}$ that predicts connections between banks, forming the graph $G = (V, E)$. Each node $v_i$ has a credit rating label $y_i \in \{1,2,3,4\}$, where lower values indicate better ratings, and these labels form the vector $Y = \{y_1, \ldots, y_N\}$. We learn $G_t(V_t, E_t)$ from $X_t$ and our task is to predict the ratings $Y_{t+1}$ of $G_{t+1}$.

\section{Our HTGNN Model}

Our Heterogeneous Topological Graph Neural Network (HTGNN) model consists of three main parts: (a) the learning of an interbank leading network using a {\it loan quota matrix}; (b) the learning of a bank network using {\it persistent homophily}; and (c) the combination of these networks for bank credit rating prediction with a {\it heterogeneous graph neural network}. Figure 2 provides a schematic representation of how the parts interact with each other.

\vspace{-0.2cm}
\begin{figure}[h]
	\[
	\begin{array}{|c|}
	\hline \\ [-11pt]
	\includegraphics[scale = 0.23]{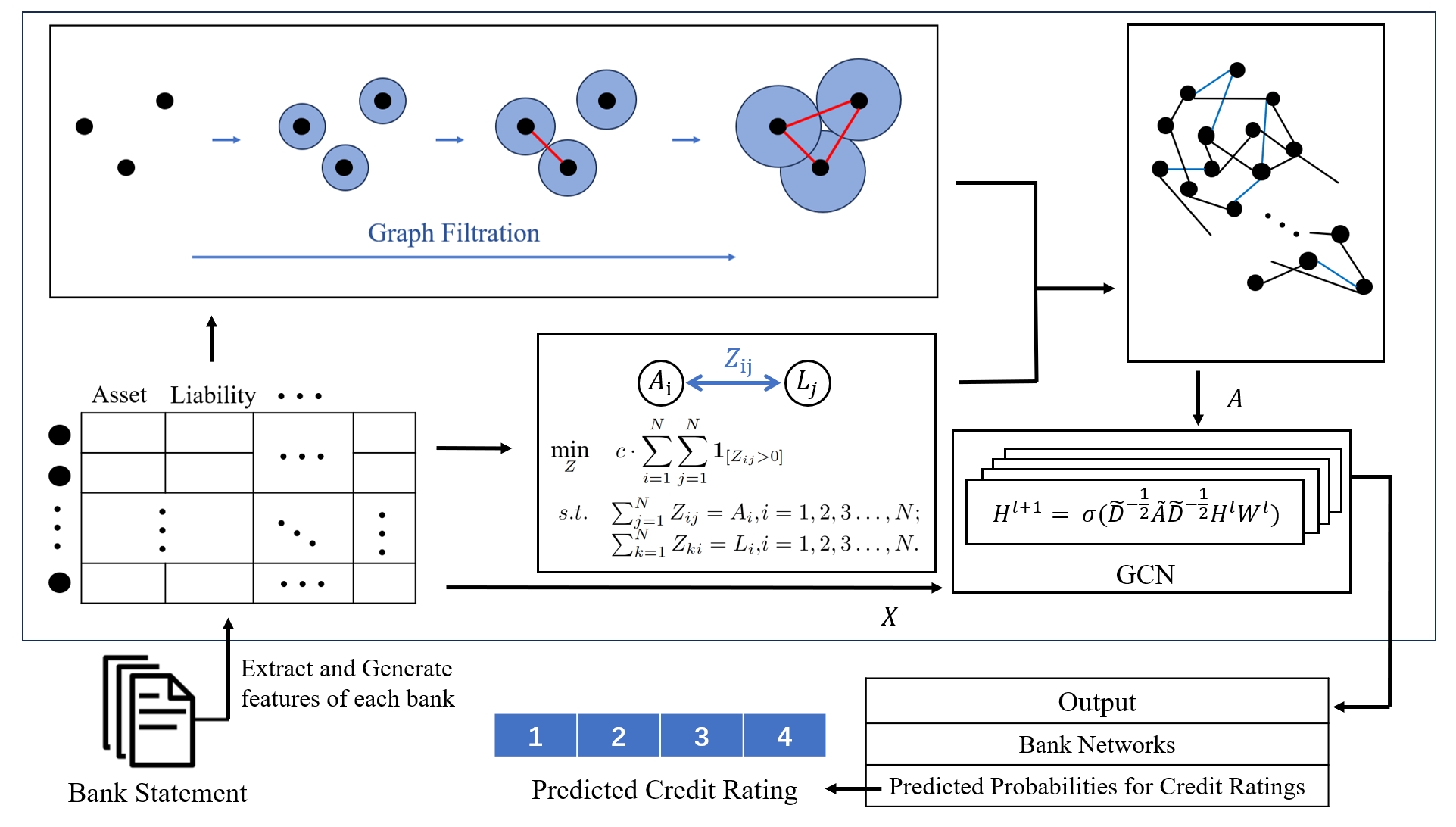} \\ [-4pt]
	\hline
	\addstackgap[7pt]{{\usefont{T1}{ptm}{b}{n}Figure 2.\hspace{0.09cm} HTGNN Model Framework.}}\\
	\hline
	\end{array}
	\]
\end{figure}
\vspace{-0.9cm}

\subsection{Loan Quota Matrix} 

Acquiring a comprehensive network of interbank lending relationships is challenging due to the commercial sensitivity of such information, which is rarely disclosed. Researchers have developed methods to construct these networks from publicly available data, like bank statements. One effective approach is the minimum density method (MDM)~\autocite{anand2015filling}. MDM assumes banks seek to minimize interbank lending relationships due to associated costs.

MDM calculates a {\it loan quota matrix} $Z$ using data from bank statements, where $Z_{ij}$ represents the loan amount from bank $i$ to bank $j$. The sums of each row and column in $Z$ correspond to bank $i$'s total loans granted and received, matching publicly reported interbank assets $A_i$ and liabilities $L_i$. By enforcing these constraints in an optimization problem, MDM infers the values of $Z$, thereby revealing the lending relationships between banks. Here, $c$ represents the fixed cost of establishing a lending relationship, and $\boldsymbol{1}_{[Z_{ij}>0]}$ is an indicator function that returns 1 if $Z_{ij}>0$.
\begin{align*} 
	&\min_{Z} \quad c \cdot \sum_{i=1}^{N} \sum_{j=1}^{N} \boldsymbol{1}_{[Z_{ij}>0]}  \\ 
	& \begin{array}{r@{\quad}l@{}l@{\quad}l} 
		s.t.& \sum_{j=1}^{N}Z_{ij} = A_{i} , 
		\sum_{k=1}^{N}Z_{ki} = L_{i}, &i=1,2,3\ldots,N. 
	\end{array} 
\end{align*} 
%
%
\subsection{Graph Homophily (or Lack Thereof)} 
%
In network analysis, graph homophily refers to the tendency of connected nodes to share similar labels or characteristics~\cite{zhu2020beyond}. For a graph \( G = (V, E) \), where \( V = \{v_1, \ldots, v_N\} \) represents nodes, \( E \) represents edges, and each node \( v_i \) has a label \( y_i \), the homophily ratio \( R \) measures the proportion of connected node pairs that share the same label. The homophily ratio is calculated using an indicator function, \( \boldsymbol{1}_{(y_i = y_j)} \), which returns 1 if the labels of nodes \( v_i \) and \( v_j \) are the same (i.e., \( y_i = y_j \)) and 0 otherwise:
    \(
    R = \frac{\sum_{(v_i,v_j) \in E} \boldsymbol{1}_{(y_i = y_j)}}{|E|}.
    \)
Low homophily complicates GNN learning because nodes rely on neighboring nodes to update their feature representations. When neighbors have different labels, it becomes challenging for GNNs to learn meaningful representations, leading to mixed signals in the features~\autocite{luan2024graph}.

\subsection{Persistent Homology} 

In our HTGNN model, we construct a bank network using persistent homology, focusing on the concepts of Rips complex, homology group, and filtration (as detailed in the Background section).

For each bank in a set of $N$ banks, we extract key financial data (e.g., liquidity, income, deposits) from its bank statement for quarter $t$ and represent it as a $d$-dimensional vector $x_i$. This vector serves as a point in $d$-dimensional space, representing bank $i$. The $N$ vectors from all banks are compiled into an $N \times d$ matrix $X=[x_1,x_2,\ldots,x_N]^T$. To create a Rips complex from these points, we define the distance between two banks $x_i$ and $x_j$ using the cosine similarity $\text{cos}(x_i,x_j)$. Specifically, the distance is calculated as \(d(x_i, x_j) = 1 - \text{cos}(x_i,x_j) = 1 - \frac{x_i \cdot x_j}{\|x_i\| \|x_j\|}\), where $x_i \cdot x_j$ is the dot product, $\|x_i\|$ is the vector length, and \(0 \leq d(x_i, x_j) \leq 1\). This metric measures the difference between the information in their respective bank statements.

Using this distance, we construct the Rips complex $G(X,r)$ by connecting pairs $(x_i,x_j)$ with an edge if their distance is within the threshold $r$. We then gradually increase $r$ from $r_0$ to $r_{\max}$, constructing a sequence of Rips complexes $G(X,r_0), G(X,r_1), \ldots, G(X,r_{\max})$. From these, we generate a filtration $H_m(G(X,r_0)) \subseteq H_m(G(X,r_1)) \subseteq \ldots \subseteq H_m(G(X,r_{\max}))$ for each homology group $H_m$, where $m \in \{0,1,2\}$. The homology group $H_m$ identifies network structures with $m$-dimensional ``holes" (connected components, loops, voids). This filtration allows us to determine the birth $r_b$ and death $r_d$ scales of each structure and compute its lifespan $r_d - r_b$. We retain structures with lifespans above a threshold $\tau$ in the set $P$.

Using the set $P$ of persistent network structures we have identified and the set of banks $\{x_1,x_2,\ldots,x_N\}$, we create a graph by linking two banks $x_i$ and $x_j$ if an edge exists between them in at least one persistent structure in $P$. We create a graph that connects banks with the robust, topological structures that we have discovered with persistent homology. We denote the graph that is learned in this manner as $G_p = (V_p, E_p)$, where $V_p = \{x_1,x_2,\ldots,x_N\}$ and $E_p$ is the set of edges that are learned. 

\subsection{Heterogeneous Graph Neural Network}

To combine the complementary insights from both graphs, $G_q = (V_q, E_q)$ learned from the loan quota matrix and $G_p = (V_p, E_p)$ learned from persistent homology, we merge them into a unified graph $G_t = (V_t, E_t)$, where $t$ denotes the quarter. Since $V_q$ and $V_p$ represent the same banks, $V_t = V_q = V_p = \{x_1,\ldots,x_N\}$, with $x_i$ as the feature vector for the $i^{th}$ bank. The edge set $E_t$ is the disjoint union of $E_q$ and $E_p$, i.e., $E_t = E_q \sqcup E_p$, with each edge annotated by its source: an edge $(x_i,x_j)$ in $E_q$ is labeled as $(x_i,x_j, q)$, and an edge $(x_m,x_n)$ in $E_p$ as $(x_m,x_n, p)$. This creates a heterogeneous graph $G_t$ paired with the credit ratings $Y_t = \{y_1, \ldots, y_N\}$.

We use a GCN (detailed in the Background section) and train it on the dataset, with modifications to accommodate the two types of edges in the unified graph. To ensure that the GCN learns characteristics unique to each edge type, we maintain separate adjacency matrices ($\tilde{A}_q,\tilde{A}_p$), normalization matrices ($\tilde{D}_q,\tilde{D}_p$), and weight matrices ($\tilde{W}_q,\tilde{W}_p$) for each edge type. The hyperparameters $\alpha_q \geq 0$ and $\alpha_p \geq 0$ weigh the relative contribution of each edge type, with $\alpha_q + \alpha_p = 1$. The propagation of information in the GCN from layer $l$ to $l+1$ is governed by the equation:
\[ H^{(l+1)} = \alpha_{q} \sigma\left( \tilde{D}_{q}^{-\frac{1}{2}} \tilde{A}_{q} \tilde{D}_{q}^{-\frac{1}{2}} H^{l} W_{q}^{l} \right) + \alpha_{p}  \sigma\left( \tilde{D}_{p}^{-\frac{1}{2}} \tilde{A}_{p} \tilde{D}_{p}^{-\frac{1}{2}} H^{l} W_{p}^{l} \right) \]
The GCN processes the unified graph through $L$ layers. The final node representations (embeddings) in the GCN, denoted by the $N \times d$ matrix $H^{(L)}$, and a learnable $d \times 4$ weight matrix $W_c$ 
are then used to predict class probabilities. First, we calculate a $N \times 4$ score matrix $Z = H^{(L)} W_c$, where $Z_{ij}$ represents the unnormalized probability that bank $i$ belongs to class $j$ (there are 4 classes, each corresponding to a credit rating). The softmax function is then applied to this score matrix to obtain a probability distribution across the four classes for each node:$ P_{ij} = \frac{e^{Z_{ij}}}{\sum_{k=1}^{4} e^{Z_{ik}}}$. The predicted class label (credit rating) $\hat{y}_i$ for bank $i$ is the class with the highest probability: $\hat{y}_i = {\arg\max}_{j} P_{ij} = {\arg\max}_{j } \frac{e^{Z_{ij}}}{\sum_{k=1}^{4} e^{Z_{ik}}}$. Next, we create a unified graph $G_{t+1} = (V_{t+1},E_{t+1})$ in the same manner as for $G_t$. The GCN is trained on $(G_t, Y_t)$ is  to predict the credit ratings $Y_{t+1}$ .

\section{Experiments}

\subsection{Dataset}


In our study, we use a global banking dataset spanning the period from the first quarter of 2019 to the last quarter of 2023, encompassing a total of 20 quarters. Sourced from the Bureau van Dijk (BVD) data platform \footnote{\url{https://orbis.bvdinfo.com/}}, this dataset incorporates information on 4548 banks worldwide. 

For each quarter, the dataset provides a vector of features for each bank extracted from the associated bank statement. Each vector contains 70 features such as equity, liquidity, long-term funding, and deposits. The credit rating per quarter for each bank is derived from existing ratings from agencies such as Moody's. The original credit ratings are divided into a dozen-plus levels, and we categorize these into four classes  $\{1,2,3,4\}$, preserving the ordinality of the original ratings (1 is the best rating and 4 is the worst). 
 
\subsection{Evaluation}

Building on the Problem Formulation section, we frame predicting bank credit ratings as a node classification task within a graph~\autocite{zhang2023bring}. Each node represents a bank, and its label corresponds to its credit rating. With four distinct rating levels, this becomes a multi-class classification challenge. For evaluating models, accuracy and F1 are common metrics. Accuracy measures the fraction of correct predictions, but can be misleading on imbalanced datasets. Precision, the fraction of true positives among predicted positives, and recall, the fraction of true positives among actual positives, offer more insight. The F1 score, the harmonic mean of precision and recall, balances these aspects: \( F1 = 2 \times \frac{\text{Precision} \times \text{Recall}}{\text{Precision} + \text{Recall}} \).
%
%
\subsection{Our Model and Baselines}

To assess the effectiveness of our HTGNN model, we benchmark it against two baselines. The first baseline, termed LQM, uses only the interbank lending network learned by applying the MDM on a loan quota matrix~\autocite{anand2015filling}. The second baseline, termed PH, uses only the bank network that is learned using persistent homology. Both baselines apply a GCN on its respective network. In contrast, our HTGNN model uses a heterogeneous graph that combines both graphs that are individually learned using MDM and PH. (all described in the HTGNN section.) These comparisons allow us to evaluate our proposed model relative to the existing network generation method (LQM), and show the effectiveness of combining graphs derived from both MDM and PH. 

The hyperparamters of our HTGNN model are set as follows:
$r_0=0, r_{max}=0.7, \alpha_0=0.1, \alpha_1=0.9$, and $\tau = 0.05$. 
The GCN in our HTGNN model is trained with the Adam optimization algorithm~\autocite{kingma2014adam} with its hyperparameters set as: $learning\_rate=0.01, epoch=1000, weight\_decay=5e-4$, and $dropout\_rate=0.5$.

\subsection{Results}

Due to GPU memory constraints, we are limited to processing 1000 banks at a time. Consequently, we randomly sample about 1000 banks from our dataset for each experiment. We train each model on the bank data from quarter $t$, and then apply the trained model to predict credit ratings for quarter $t+1$. To ensure robustness, we ran each experiment 10 times and report the average results. 
\bgroup
\def\arraystretch{1.2}
\footnotesize
\begin{table}[ht]
    \centering
    \begin{tabular}{|c|c|c|c|c|c|c|c|c|}
        \hline
         
        & \multirow{2}{*}{ \textbf{Model}} &\multirow{2}{*}{ \textbf{LMQ}} & \multirow{2}{*}{\textbf{PH}} & \multirow{2}{*}{\textbf{HTGNN}} &\multirow{2}{*}{ \textbf{LMQ}} & \multirow{2}{*}{\textbf{PH}} & \multirow{2}{*}{\textbf{HTGNN}} \\
           &    &   & \textit{(ours)} & \textit{(ours)} &      & \textit{(ours)} & \textit{(ours)} \\
    \hline
         & \textbf{Year} 
        & \multicolumn{3}{c|}{\textbf{2019}} & \multicolumn{3}{c|}{\textbf{2020}} \\ \hline
        \multirow{5}{*}{\textbf{Metric}}  & Accuracy & 51.55 & 55.21 & \textbf{56.82} & 57.51 & 62.57 & \textbf{63.06}  \\  \cline{2-8}
        &F1 & 42.62&44.85 & \textbf{47.56} & 47.42& 56.52 & \textbf{56.70}  \\  \cline{2-8} 
        & Precision & 51.68& 61.72 & \textbf{64.20}&71.02 & 70.82 & \textbf{72.20}  \\  \cline{2-8}
         &Recall & 51.16& 55.70 & \textbf{56.88} &48.19 & 57.50 & \textbf{58.21}  \\ \hline
        & \textbf{Year} 
        & \multicolumn{3}{c|}{\textbf{2021}} & \multicolumn{3}{c|}{\textbf{2022}} \\ \hline
        \multirow{5}{*}{\textbf{Metric}}  & Accuracy & 61.68 & 65.67 & \textbf{67.02} & 66.22 &  69.63 & \textbf{70.57}  \\  \cline{2-8}
        &F1 &49.26 &54.01 & \textbf{54.96} & 54.70 &56.31& \textbf{56.60}  \\  \cline{2-8} 
        & Precision & 70.41& 74.61 & \textbf{74.84} & 68.37 & \textbf{69.87}  &  69.79\\  \cline{2-8}
         &Recall & 50.93 & 56.85& \textbf{57.33} & 56.80 & 59.99 & \textbf{60.07}   \\ \hline
        & \textbf{Year} 
        & \multicolumn{3}{c|}{\textbf{2023}} &  \multicolumn{3}{c|}{\textbf{}} \\ \hline
        \multirow{5}{*}{\textbf{Metric}}  & Accuracy & 56.83 & 61.05 & \textbf{62.34} &  &  & \textbf{}  \\  \cline{2-8}
        &F1 & 51.08 & 55.73 & \textbf{56.68} &  &  & \textbf{}  \\  \cline{2-8} 
        & Precision & 67.98 & 69.17 & \textbf{69.18} &  &  & \textbf{}  \\  \cline{2-8}
         &Recall & 52.19 & 58.27 & \textbf{58.59} &  &  & \textbf{}  \\ \hline
        \multicolumn{8}{|c|}{\normalsize{{\usefont{T1}{ptm}{b}{n}Table 1. \hspace{0.09cm} Accuracy, F1, Precision, and Recall (as Percentages) from 2019 to 2023.}}} \rule{0pt}{3ex} \\ [4pt] \hline
    \end{tabular}
\end{table}
\egroup
\vspace{-0.5cm}
Table 1 summarizes the performance of our model compared to the baselines, with metrics averaged across all four quarters of each year from 2019 to 2023 (due to page constraints, quarterly results are not listed). For clarity, the best-performing model for each metric is highlighted in bold. Our HTGNN model consistently outperforms both baselines. Specifically, while the PH baseline—leveraging persistent homology—consistently surpasses the LMQ baseline, our HTGNN model shows an approximately 2\% higher accuracy than PH on average. In contrast, LMQ, which relies on the minimum density method, trails behind HTGNN by about 6\% in accuracy. Although HTGNN slightly underperforms PH in precision for one year, it achieves a better balance between precision and recall, as evidenced by its superior F1 scores. This suggests that HTGNN offers a more balanced approach, excelling in both recall and precision.

To assess the statistical significance of these performance differences, we conduct paired t-tests. Table 2 presents the t-statistic, p-value, and significance level for each comparison. The results show that PH's superiority over LMQ is statistically significant, as is HTGNN's superiority over PH. Overall, HTGNN demonstrates statistically significant improvements over both baselines.
\bgroup
\def\arraystretch{1.2}
\footnotesize
\begin{table}[h]
    \centering
    \begin{tabular}{|c|c|c|c|}
        \hline
        \textbf{Model Comparison} & \textbf{t-Statistic} & \textbf{ p-Value} & \textbf{Statistical Significance} \\ \hline
        LMQ vs PH & -7.6021 & 3.5457e-07 & ** \\ \hline
        LMQ vs HTGNN & -8.1192  & 1.3445e-07 & ** \\ \hline
        PH vs HTGNN & -5.0261 & 7.4999e-05 & ** \\ \hline
                \multicolumn{4}{|c|}{\normalsize{{\usefont{T1}{ptm}{b}{n}Table 2. \hspace{0.09cm} Paired t-Test Comparisons of Model Performances. (** respectively indicate}}} \rule{0pt}{3ex} \\ [1pt] 
        \multicolumn{4}{|c|}{\normalsize{{\usefont{T1}{ptm}{b}{n} $p<0.01$; negative indicates the first model performs lower than the second. }}} \rule{0pt}{3ex} \\ [1pt] 
        \hline
    \end{tabular}
\smallskip  
\noindent  
\end{table}
\egroup
\vspace{-0.5cm}
\section{Conclusion and Future Work}



We propose the Heterogeneous Topological Graph Neural Network (HTGNN) model, designed to learn bank networks for predicting credit ratings. We identify the lack of homophily in interbank lending networks learned by state-of-the-art methods and demonstrate how this hinders tasks like credit rating prediction. To address this, we introduce persistent homology to learn networks that preserve robust structural regularities and exhibit higher homophily, aiding in accurate credit rating predictions. Our empirical experiments with real-world banking data show the effectiveness of our approach compared to two competitive baselines. As future work, we plan to extend HTGNN to detect risk contagion in financial networks, an area that remains relatively underexplored.

\printbibliography   

\end{document}